\documentclass[10pt,twocolumn,letterpaper]{article}

\usepackage{cvpr}              

\usepackage{graphicx}
\usepackage{amsmath}
\usepackage{amssymb}
\usepackage{booktabs}
\usepackage{multirow}
\usepackage[ruled,linesnumbered]{algorithm2e} 
\usepackage{xcolor}
\usepackage[pagebackref,breaklinks,colorlinks]{hyperref}

\usepackage[capitalize]{cleveref}
\crefname{section}{Sec.}{Secs.}
\Crefname{section}{Section}{Sections}
\Crefname{table}{Table}{Tables}
\crefname{table}{Tab.}{Tabs.}

\def\cvprPaperID{8714} 
\def\confName{CVPR}
\def\confYear{2023}

\begin{document}

\title{Motion-R\textsuperscript{3}: Fast and Accurate Motion Annotation via Representation-based Representativeness Ranking}

\author{Jubo Yu\\
Xiamen University\\
{\tt\small 30920201153951@stu.xmu.edu.cn}
\and
Tianxiang Ren\\
Xiamen University\\
{\tt\small 30920201153939@stu.xmu.edu.cn}
\and
Shihui Guo\\
Xiamen University\\
{\tt\small guoshihui@xmu.edu.cn}
\and
Fengyi Fang\\
Xiamen University\\
{\tt\small 34520182201469@stu.xmu.edu.cn}
\and
Kai Wang\\
Nanjing University Of Aeronautics and Astronautics\\
{\tt\small wk0806@nuaa.edu.cn}
\and
Zijiao Zeng\\
Tencent Technology\\
{\tt\small zijiaozeng@tencent.com}
\and
Yazhan Zhang\\
Tencent Technology\\
{\tt\small yazhanzhang@tencent.com}
\and
Andreas Aristidou\\
University of Cyprus\\
{\tt\small a.m.aristidou@gmail.com} 
\and
Yipeng Qin\\
Cardiff University\\
{\tt\small qiny16@cardiff.ac.uk}
}

\twocolumn[{
\renewcommand\twocolumn[1][]{#1}
\maketitle
\begin{center}
    \captionsetup{type=figure}
    \includegraphics[width=\textwidth]{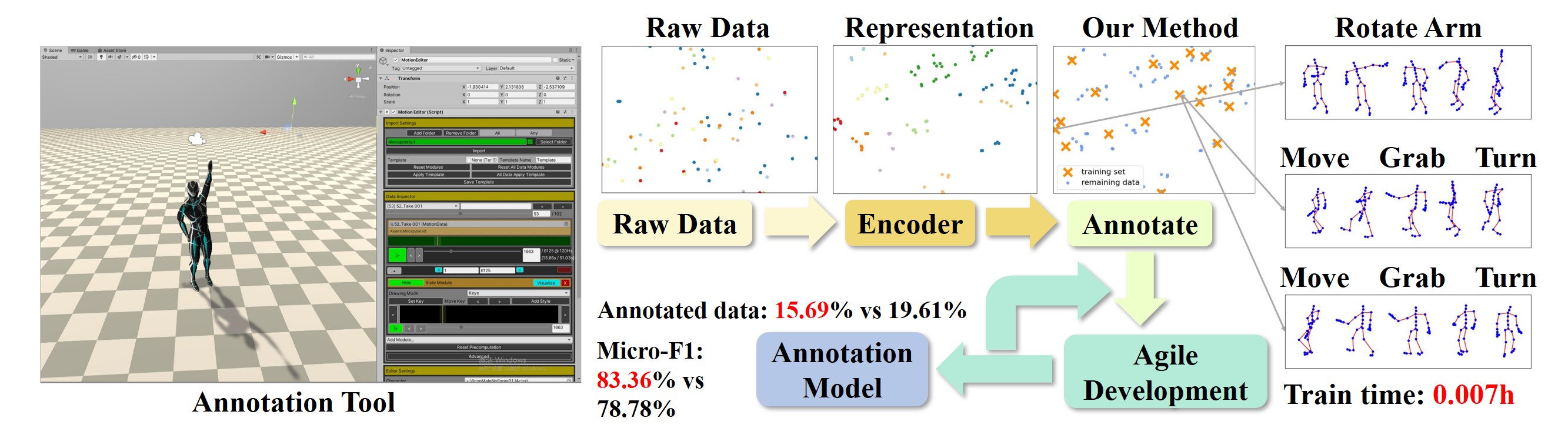}
    \caption{Overview of the proposed Motion-R\textsuperscript{3} method. Our method (bottom) i) reduces the time cost to (re)train motion annotation models by a factor of 500 to 700+, which allows for agile development; ii) achieves better micro-F1 scores with less manually annotated data, reducing the labour and time cost for manual annotation.
Please see the {\it supplementary materials} for the demo of the proposed method.}
\end{center}
}]

\begin{abstract}
In this paper, we follow a {\it data-centric} philosophy and propose a novel motion annotation method based on the inherent representativeness of motion data in a given dataset.
Specifically, we propose a Representation-based Representativeness Ranking (R\textsuperscript{3}) method that ranks all motion data in a given dataset according to their representativeness in a learned motion representation space. 
We further propose a novel dual-level motion constrastive learning method to learn the motion representation space in a more informative way.
Thanks to its high efficiency, our method is particularly responsive to frequent requirements change and enables agile development of motion annotation models.
Experimental results on the HDM05 dataset against state-of-the-art methods demonstrate the superiority of our method. 

\end{abstract}

\section{Introduction}

Along with the recent AI boom, data driven character animation has been revolutionized and dominated by deep learning~\cite{starke2021neural, starke2020local}.
Despite its success, deep learning is known to be data-hungry, which poses challenges for both academia and industry as high-quality annotated data are usually expensive and difficult to obtain.
This is even more challenging for mocap (motion capture) data due to the large amount of data frames obtained from dense captures and the complex annotation procedure where multiple labels could be assigned to a single frame (i.e., an actor may wave while walking).

To minimize labour costs in annotation tasks, the best-performing methods resort to machine learning solutions.
For example, Müller et al.~\cite{muller2009efficient} proposed to use motion templates and dynamic time warping (DTW) distance to segment and annotate motion data; Carrara et al.~\cite{carrara2019lstm} proposed to use long short-term memory (LSTM) network to predict motion labels.
Despite their differences, all these methods are {\it model-centric} and trained with expert-picked training data that are not suitable for machine use.


In this paper, we follow the {\it data-centric AI} philosophy advocated by Andrew Ng~\cite{ng2022unbiggen} and argue that the performance of motion annotation models can be significantly improved by simply using more representative samples in the training.
Specifically, inspired by the classic farthest point sampling strategy, we propose a Representation-based Representativeness Ranking (R\textsuperscript{3}) method that ranks all motion data in a given dataset according to their ``representativeness'' in a learned motion representation space $\mathcal{R}$.
To learn a more informative $\mathcal{R}$, we propose a novel dual-level contrastive learning method applied on both motion sequence and frame levels.
In addition, the motion representation space $\mathcal{R}$ learned by our method is independent to specific motion annotation tasks.
This suggests that it is born to be adaptive to various motion annotation tasks, making it more responsive to frequent requirement changes and enabling agile development of motion annotation models.
Experimental results on the HDM05 dataset against state-of-the-art methods demonstrate the superiority of our method. 
In summary, our main contributions include:
\begin{itemize}
    \item We propose a novel motion annotation method (Motion-R\textsuperscript{3}) which significantly reduces the manual annotation workload without sacrificing the accuracy.
    Our method can rank motion data according to their ``representativeness''. Results show that automatic motion annotation benefits significantly from the use of more representative training samples.
    \item We propose a novel dual-level motion contrastive learning method that can learn a more informative representation space for motion data.
    \item Our method only relies on the motion representations learned in an unsupervised way, which is more responsive to frequent requirement changes and enables agile development of motion annotation models.
\end{itemize}
\section{Related Work}

\subsection{Motion Annotation}

Motion annotation aims to annotate raw and unsegmented motion data with action labels, which is a complex and tedious task as multiple action labels can be assigned to the same piece of data~\cite{bernard2013motionexplorer,zhou2012hierarchical}.
To address its challenges, a straightforward idea is to first divide the raw mocap data into action segments and then classify them respectively.
For example, the sliding window method was employed to divide raw mocap data into overlapping action segments~\cite{muller2009efficient, meshry2016linear, wu2017recognition, xu2017learning} or non-overlapping semantic segments~\cite{papadopoulos2019two, boulahia2018cudi3d, devanne2017motion}.
The classification of segmented action segments is usually referred to as an action recognition task, which aims to classify each action segment to the correct action category across different spatio-temporal configurations ({\it e.g.,} velocity, temporal or spatial location)~\cite{chen2017survey,sedmidubsky2018effective, zhu2016co, du2015hierarchical, evangelidis2014skeletal, kadu2014automatic, barnachon2014ongoing, liu2017skeleton, nunez2018convolutional, song2018spatio}.
Compared to traditional model-based methods~\cite{xia2012view, wang2013approach, kadu2014automatic} and classifiers \cite{raptis2011real, zanfir2013moving, vieira2012distance}, state-of-the-art action recognition methods resort to deep convolutional neural networks \cite{laraba20173d, sedmidubsky2018effective} and LSTM neural networks \cite{zhu2016co, nunez2018convolutional, singh2017human}
to effectively model spatial and temporal motion features, as deep learning has demonstrated its power in identifying complex patterns in multimedia data \cite{asadi2017survey, baltruvsaitis2018multimodal}.
On the other hand, frame-based motion annotation methods have recently gained popularity as they are more fine-grained and can predict the probabilities of each action per frame directly. 
The classification tasks in these methods are usually implemented by vector machines~\cite{sharaf2015real}, linear classifiers~\cite{zhao2013online}, structured streaming skeletons~\cite{zhao2014structured}, LSTM networks~\cite{li2016online, song2018spatio, carrara2019lstm}, etc.
Furthermore, flow-based methods can identify motion before the motor behavior ends~\cite{li2018early} and even predict future action~\cite{jain2016structural, wu2015watch}.

In this work, we investigate an important  but under-explored problem in motion annotation, {\it i.e.,} the representativeness of mocap data points.
We demonstrate that the performance of motion annotation can be significantly improved by simply picking more representative samples for training ({\it i.e.,} Representation-based Representativeness Ranking), which is orthogonal to all existing works.

\begin{figure*}[!ht]
  \includegraphics[width=\textwidth]{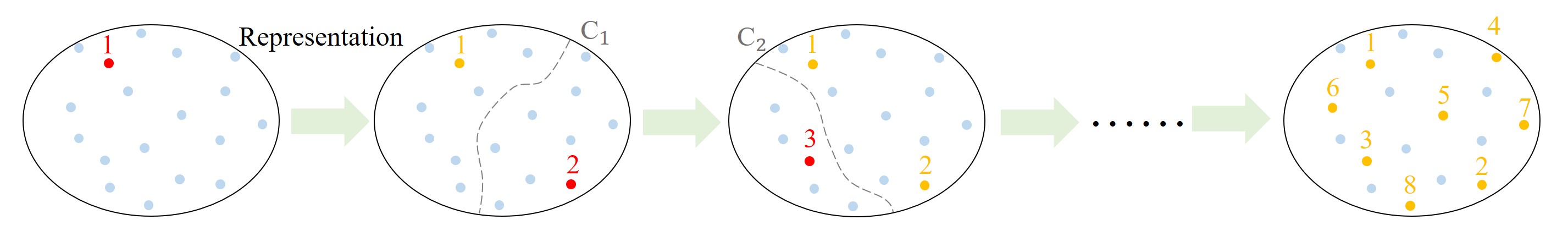}
\caption{Illustration of our Representation-based Representativeness Ranking (R\textsuperscript{3}) method.
\textcolor{blue}{Blue}: motion sequences to be ranked.
\textcolor{orange}{Orange}: ranked motion sequences. \textcolor{red}{Red}: the next motion sequence in the ranking. $C_n$: a binary classifier.}
  \label{fig:alp}
\end{figure*}

\subsection{Contrastive Learning}
Contrastive learning is an unsupervised representation learning method that can learn high-quality feature spaces from unlabeled data \cite{tian2020contrastive, van2018representation, wu2018unsupervised, he2020momentum, chen2020simple, chen2020improved}.
Contrastive Learning has made great progress in the field of computer vision~\cite{vincent2008extracting,zhang2016colorful,hadsell2006dimensionality,wu2018unsupervised,gutmann2010noise,tian2020contrastive,caron2020unsupervised,chen2020simple,he2020momentum, chen2020improved,chen2020big,chen2021exploring}. And Momentum Contrastive Paradigm (MoCo) \cite{he2020momentum, chen2020improved} facilitates contrastive unsupervised learning through a queue-based dictionary lookup mechanism and momentum-based updates.

Contrastive learning has already been applied and achieved promising results in motion-related tasks.
MS2L \cite{lin2020ms2l} integrates contrastive learning into a multi-task learning framework; AS-CAL \cite{rao2021augmented} uses different backbone sequence augmentations to generate positive and negative pairs; Thoker et al. \cite{thoker2021skeleton} perform representation learning in a graph-based and sequence-based mode using two different network architectures in a cross-contrasted manner.
Recently, SkeletonCLR \cite{li20213d} learns skeleton sequence representations through a momentum contrast framework.
In a concurrent work, AimCLR \cite{guo2021contrastive} extends SkeletonCLR with an energy-based attention-guided casting module and nearest neighbor mining.
BYOL~\cite{moliner2022bootstrapped} extends representation learning for skeleton sequence data and proposes a new data augmentation strategy, including two asymmetric transformation pipelines.

In this work, we propose a novel {\it dual-level} motion contrastive learning approach which extends MoCo~\cite{he2020momentum, chen2020improved} to motion data and implements contrastive learning at both sequence and frame levels, which works as the basis of the proposed Motion-R\textsuperscript{3} method.

\subsection{Motion Represent}
For example,Aristidou\cite{aristidou2018deep} employs comparative learning to generate high-dimensional motion features, which can be used in many applications for indexing, temporal segmentation, retrieval, and synthesis of motion clips.
Bernard\cite{bernard2013motionexplorer} operates a combination of hierarchical algorithms to create search groups and extract motion sequences.
Zhou\cite{zhou2012hierarchical} applies alignment clustering analysis to action segmentation and expands standard kernel kmeans clustering through dynamic time warping (DTW) kernel to achieve temporary variance.
Forbes\cite{forbes2005efficient} chooses weighted PCA to represent the pose, combined with the calculation of pose-to-pose distance, which is flexible and efficient, searching for similar motion sequences.
Holden\cite{holden2015learning} utilizes an automatic convolutional encoder to learn a variety of human motion manifolds as a motion priori to resolve ambiguity.

\section{Our Motion-R\textsuperscript{3} Method}

Let $D=(x_{1},x_{2}, ..., x_{N})$ be a motion dataset consisting of $N$ motion sequences, $x_i = (s_{i,1},s_{i,2}, ..., s_{i,T})$ be a motion sequence consisting of $T$ consecutive skeleton pose frames, $s_{i,j} \in \mathbb{R}^{J \times 3}$ ($j=1,2,...,T$) be the 3D coordinates of the $J$ body joints of a skeleton pose, we aim to assign a binary label vector $\mathbf{c_{i,j}} = \{0, 1\}^m$ to each skeleton pose $s_{i,j}$ where ${c_{i,j,k}}=1$ ($k=1,2,...,m$) if $s_{i,j}$ belongs to the $k$-th class of $m$ pre-defined motion types.
To minimize labour costs, we assume only a small portion of $D$ are manually annotated as $D_{\mathrm{train}}$ and the rest can be automatically annotated by a machine learning model trained with $D_{\mathrm{train}}$ as the training set.

\subsection{Overview}



Unlike previous methods~\cite{carrara2019lstm} which select $D_{\mathrm{train}}$ by visual inspection, we argue that picking the more {\it representative} ones for manual annotation not only reduces the labour and time costs but also increases the model's accuracy.
As Fig.~\ref{fig:alp} shows, our method aims to learn a representativeness ranking of motion sequences $x_i \in D$ in an unsupervised manner using the inherent similarities and differences among them by classifier $C$ which consists of three linear layers.
Alg.~\ref{alg:recommendation} shows the pseudo-code of our R\textsuperscript{3} method.

\paragraph{Representation Learning}
We first train a feature encoder $E$ which learns a representation space $\mathcal{R}$ for $x_i $ in an unsupervised manner.
For the learning of feature encoder $E$ and motion representation space $\mathcal{R}$, we adopt one of the latest contrastive learning approach: Momentum Contrast (MoCo)~\cite{he2020momentum, chen2020improved}, which has recently demonstrated superior performance and generalization abilities in computer vision tasks.
Nevertheless, MoCo was designed for computer vision tasks and only works on the 2D grid-like image data.
Thus, it is non-trivial to acclimatize it to motion sequences.
Addressing this issue, we propose a {\it dual-level} motion contrastive learning approach which extends MoCo to motion data and implements contrastive learning at both sequence and frame levels, which is depicted in the next subsection.

\begin{algorithm}[t]
\caption{R\textsuperscript{3}: Representation-based Representative Ranking}
\label{alg:recommendation}
\KwData{Motion representation space $\mathcal{R}$, motion dataset $D = [x_{1},x_{2}, ..., x_{T}]$ and $\hat{D}=[]$, binary classifier $C$.}
\KwResult{Sorted motion dataset $\hat{D}$.}
$a \gets$ a random number in $\{1,2,...,T\}$\; 
$\hat{D} \gets \hat{D}$.append($x_a$), $D \gets D$.remove($x_a$)\;
\While{$D \neq []$}{
    Train $C$ to distinguish between elements of $\hat{D}$ and $D$ in $\mathcal{R}$\;
    $x_a \gets$ the most representative $x_i \in D$ according to $C$\;
    $\hat{D} \gets \hat{D}$.append($x_a$), $D \gets D$.remove($x_a$)\;
}
\end{algorithm}

\paragraph{Representativeness Ranking}
We next rank $x_i \in D$ according to their representativeness in $\mathcal{R}$.
Inspired by the classic farthest point sampling strategy, we implement our R\textsuperscript{3} method by progressively including $x_i \in D$ to a sorted motion dataset $\hat{D}$, where $x_i$ is the ``farthest'' ({\it i.e.,} the most representative) motion sequence to the ranked ones in $\hat{D}$ in the representation space $\mathcal{R}$.
The ranking is then determined by the order in which $x_i$ is included into $\hat{D}$.
Note that we search the ``farthest`` points via binary classification to avoid the high computational costs of traditional farthest point sampling methods that consumes $O(n^2)$ time with decreasing to $O(n)$.

\paragraph{Motion Annotation with R\textsuperscript{3}}
For motion annotation, we first assign motion sequences to human annotators according to the ranking $\hat{D}$ and get $\hat{D}_{\mathrm{train}}$. 
Then, we train a low-cost and simple classifier $C_{\mathrm{simple}}$ using the learned representation $\mathcal{R}$ and $\hat{D}_{\mathrm{train}}$ to annotate the remaining motion sequences automatically.

\subsection{Dual-level Motion Contrastive Learning}
\begin{figure}
\begin{minipage}{\columnwidth}
  \includegraphics[width=\linewidth]{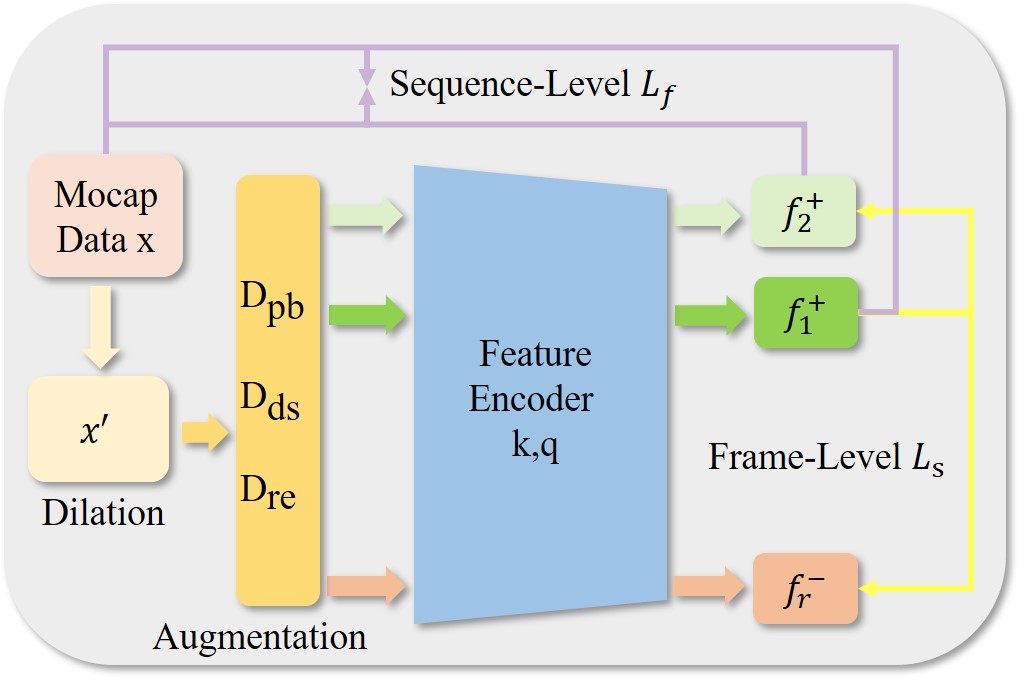}
  \caption{Dual-level Motion Contrastive Learning.}
  \label{fig:clp}
\end{minipage}
\end{figure}

In a nutshell, contrastive learning assumes that a good data representation has two properties: similar data points should be close to each other in the feature space, while different data points should be far from each other.
Accordingly, it proposes to fulfil the two properties by minimizing the distances among positively augmented samples and maximizing those among negatively augmented samples.
Building on this idea, MoCo~\cite{he2020momentum, chen2020improved} shows that the performance of contrastive learning can be boosted by maintaining a large and consistent dictionary of negatively augmented samples, which is implemented by the incorporation of a queue and a momentum encoder.
Thus, the extension of MoCo to {\it motion data} boils down to three questions: i) how to select a proper backbone network for feature encoding? ii) how to design the positive and negative data augmentation methods? iii) how to measure the distances between samples in the feature spaces ({\it i.e.,} the contrastive loss)?

\subsubsection{Dilated (Momentum) Feature Encoder.}
Since motion data are usually captured at a high sampling rate ({\it e.g.,} 120 FPS), the differences between adjacent frames are tiny, which causes ambiguities that confuse the model in identifying the action of a single frame.
To clarify such ambiguity, we borrow the idea of {\it dilated} convolution~\cite{Yu2016DilatedConvolution} and enhance each input frame with its context information ({\it i.e.,} dilated joint trajectory) in a time window $t$ centered at the current frame. 
Specifically, assuming the sampling rate is $r=120$ FPS, we employ a dilution factor $l$ that enhances input frame $s_{i,j}$ with its context information as 
\begin{equation}
    \begin{split}
        s'_{i,j} = (s_{i,j} - s_{i,j-nl}, ...,  s_{i,j} - s_{i,j-l}, \\
                    s_{i,j}, s_{i,j+l} - s_{i,j}, ...,  s_{i,j+nl} - s_{i,j})
    \end{split}
    \label{eq:dilated_input}
\end{equation}
where $n = \lfloor t \cdot r/l \rfloor$, $\lfloor \cdot \rfloor$ is a flooring function,  $\pm(s_{i,j} - s_{i, j+kl})$ denotes the dilated joint trajectory, $k = \{-n, -n+1, ... n\}$.
We use $x'_i=(s'_{i,1},s'_{i,2}, ..., s'_{i,T})$ as the input to our (momentum) feature encoders.
We replace the Vision Transformer~\cite{dosovitskiy2020image} with similar method as Spatial Temporal Transformer ~\cite{plizzari2021spatial} as our feature encoder, for its success in modeling the dependencies among skeleton joints.
Specifically, after being embedded in a two-layer MLP network, its models the relationships among joints of a single skeleton in each frame with the so-called Spatial Transformer ($E_{ST}$) module and those among the same joints across different frames in $x'_i$ with its Temporal Self-Transformer ($E_{TT}$) module. 

For the momentum feature encoder, we follow MoCo~\cite{he2020momentum, chen2020improved} and update its parameters by:
\begin{equation}
    \theta_k \leftarrow \alpha \theta_k + (1-\alpha) \theta_q 
\end{equation}
where $\theta_k, \theta_q$ denote the parameters of the momentum and native encoders, $\alpha \in [0,1)$ denotes a momentum coefficient.

Since our motion data is a motion {\it sequence} consisting of consecutive {\it frames} of skeleton poses, we propose to implement contrastive learning at both levels as follows.

\begin{figure*}[!ht]
  \includegraphics[width=\linewidth]{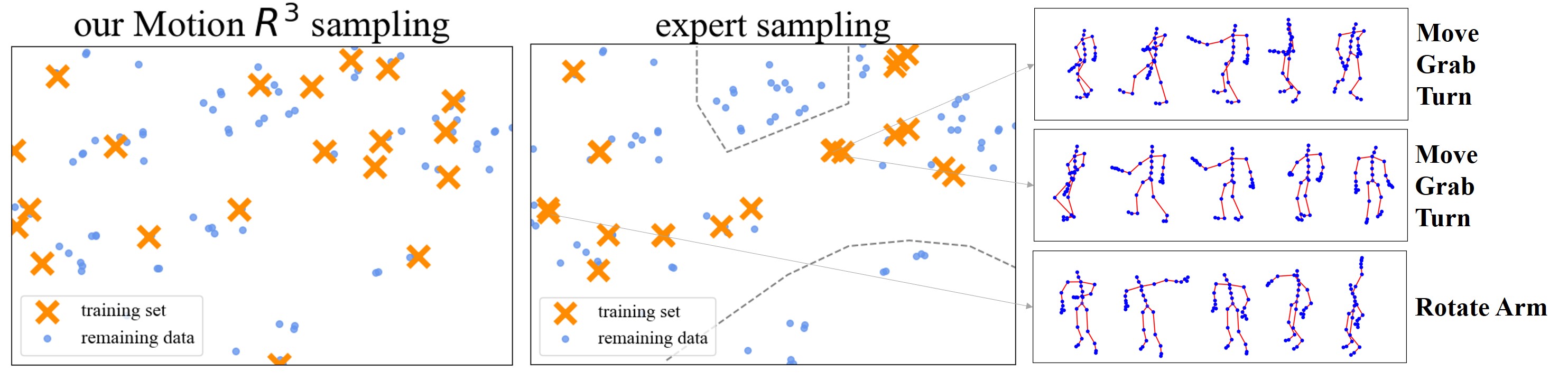}
  \caption{Comparison of sample representativeness of our Motion-R\textsuperscript{3} method against expert selection~\cite{carrara2019lstm} in the feature space.}
 \label{fig:csm}
\end{figure*}

\subsubsection{Sequence-Level Contrastive Learning} 
We show the data augmentation methods and loss design of the proposed sequence-level contrastive learning method below.

\paragraph{Sequence-level Data Augmentation.}
Similar to the applications of contrastive learning in computer vision tasks~\cite{he2020momentum, chen2020simple, chen2020improved}, the key challenge of motion data augmentation is to disentangle the inherent patterns of a skeleton from its different {\it views}, {\it i.e.,} the different appearances of the same pattern.
Addressing this challenge, let $x' = (s'_1, s'_2, ..., s'_T)$ be the enhanced input (Eq.~\ref{eq:dilated_input}),

\vspace{2mm}
\noindent i) We propose a {\it perturbation} data augmentation strategy with the rationale that motion semantics are robust against small perturbations.
Specifically, we apply two stochastic perturbations, data missing and disorder, to each input frame $s'_i$ according to $p_i \sim \mathcal{U}[0,1]$ as follows:
\begin{equation}
\begin{split}
    pb(s'_{i},p_i)=\left\{  
             \begin{array}{lr}  
             0, &  p_{i}<t_{pb}\cdot t_{md}\\  
             s'_{j}, & t_{pb}\cdot t_{md}\leq p_{i}<t_{pb}
             \end{array}  
    \right., j \sim \mathcal{U}\{1,T\}
\end{split}
\label{eq:perturbation}
\end{equation}
where $t_{pb}=0.15\in [0,1]$ is the probability threshold that $s'_i$ is perturbed, $t_{md}=0.9\in [0,1]$ is the probability threshold that missing data perturbation is applied, $t_{pb}\cdot t_{md}$ means that $s'_i$ is perturbed with missing data perturbation, the replacement of $s'_i$ with $s'_j$ denotes the disorder perturbation, $\mathcal{U}\{1,T\}$ denotes a discrete Uniform distribution from 1 to $T$.
Let $p = {p_i}$, we have
\begin{equation}
\begin{split}
    \mathcal{D}_{pb}(x',p)
    = (pb({s}'_{1},p_1),pb({s}'_{2},p_2),...,pb({s}'_{T},p_T))
\end{split}
\end{equation}

\noindent ii) Inspired by the fact that human beings can successfully recognize motions at different playback speeds ({\it i.e.,} the motion semantics are largely independent of the playback speeds), we propose a novel {\it downsampling} data augmentation technique that creates novel views of motion data by downsampling them at random rates and offsets:
\begin{equation}
\begin{split}
    \mathcal{D}_{ds}(x', a, \delta) = ({s}'_{a},{s}'_{a+\delta}, {s}'_{a+2\delta},...,{s}'_{a+(n_{ds}-1)\delta})
\end{split}
\label{eq:downsampling}
\end{equation}
where $a$ denotes the offset, $\delta$ denotes the downsampling interval, $n_{ds}=512$ denotes the number of resulting samples. Note that $a+(n_{ds}-1)\delta \leq T$.

\noindent iii) We also propose the {\it reverse} augmentation that works as a negative augmentation method:
\begin{equation}
    \mathcal{D}_{re}(x') = ({s}'_{T},{s}'_{T-1},...,{s}'_{1})
\end{equation}
With the aforementioned data augmentation methods, we generate two positively augmented views $v_1^+, v_2^+$ and a negatively augmented view $v_r^-$ as follows:
\begin{equation}
\begin{split}
    v_1^+ &= \mathcal{D}_{pb}(\mathcal{D}_{ds}(x', a^1,\delta^1), p^1)\\
    v_2^+ &= \mathcal{D}_{pb}(\mathcal{D}_{ds}(x', a^2,\delta^2), p^2)\\
    v_r^- &= \mathcal{D}_{re}(\mathcal{D}_{mk}(\mathcal{D}_{ds}(x', a^3,\delta^3), p^3))
\end{split}
\end{equation}
where $a^i, \delta^i, p^i$ denote different parameters generated randomly.

We encode these augmented views and get their normalized features:
\begin{equation}
    f_1^+ = \frac{E(v_1^+)}{\|E(v_1^+)\|}, f_2^+ = \frac{E(v_2^+)}{\|E(v_2^+)\|}, 
    f_r^- = \frac{E(v_r^-)}{\|E(v_r^-)\|}
\end{equation}
where $E$ is the dilated (momentum) feature encoder.

\paragraph{Sequence-level Contrastive Loss.}
We design our loss function based on an InfoNCE loss:
\begin{equation}
\label{eqn:tcl}
\mathcal{L}_{s}= -\log_{}{\frac{\exp (f_{1}^+ \cdot f_{2}^+/\tau)}{\exp (f_{1}^+ \cdot f_{2}^+/\tau)+ {\textstyle \sum_{i=1}^{K} \exp (f_{1}^+ \cdot f_i^- /\tau)}  } } 
\end{equation}
where $\cdot$ denotes the measurement of cosine similarity, $\tau$ is   a temperature softening hyper-parameter and $i$ denotes the indices of the negative samples $f_i^-$ maintained in the queue $Q$ of size $K$ that
\begin{equation}
    Q = {f_r^-}^\frown\{f_1^-,f_2^-\}
\end{equation}
where $^\frown$ denotes the enqueue operation, $\{f_1^-,f_2^-\}$ denotes the positive samples generated previously but are used as negative samples for $f_1^+$ as they are generated from different $x'$.

\subsubsection{Frame-level Contrastive Learning}
\label{sec:frame-level}

\paragraph{Frame-level Data Augmentation.}
Leveraging the {\it local consistency} among consecutive frames in a motion sequence ({\it i.e.,} the actions in a small neighbourhood share similar motion semantics), for the feature $f^+_{1,i}$ of each frame $s'_i$, we define $f^+_{2,j}$ are its {\it positive} samples if $j\in \Omega_{+}$ and $\Omega_{+}=\{j | t_{nb} > \left|i-j\right| \}$, where $t_{nb}=12$ is the size of the neighbourhood, and vice versa ($f^-_{2,j}$ is {\it negative} if $j\in \Omega_{-}$ and $\Omega_{-}=\{j | t_{nb} \leq \left|i-j\right| \}$). 


\paragraph{Frame-level Contrastive Loss.} Accordingly, we design our frame-level local consistency loss as:
\begin{equation}
\label{eqn:fcl}
\mathcal{L}_{\text {f}}=-\log \sum_i \sum\limits_{j \in \Omega_{+}} \exp{ \left(f^+_{1,i} \cdot f_{2,j}^{+}/ \tau \right) }
\end{equation}

\subsubsection{Overall Loss Function}

Combining the sequence-level loss $\mathcal{L}_{s}$ and the frame-level loss $\mathcal{L}_{\text {f}}$, we have:
\begin{equation}
\label{eq:overall_loss}
\mathcal{L}=\mathcal{L}_{s}+\omega \mathcal{L}_{\text {f}}
\end{equation}
where $\omega=1$ is a weighting parameter.

\section{Experiments \& Results}


\subsection{Implementation Details}
We conduct experiments on a PC with an Intel i7-7700 CPU and a Nvidea TESLA P40 GPU.
We implement our method with PyTorch.
We follow the method of Sedmidubsky J\cite{sedmidubsky2018effective} to process motion data, and normalize the position, orientation and Skeleton size. Following~\cite{carrara2019lstm}, we evaluate our method on the three variants of the HDM05 dataset~\cite{muller2007documentation}:
\begin{itemize}
    \item {\bf HDM05-15}: 102 motion sequences in 15 classes, consuming 68 minutes and 491,847 frames;
    \item {\bf HDM05-65}: 2,345 motion sequences in 65 classes, consuming 156 minutes and 1,125,652 frames;
    \item {\bf HDM05-122}: 2,238 motion sequences in 122 classes, consuming 156 minutes and 1,125,652 frames.
\end{itemize}
Note that HDM05-65 and HDM05-122 contain the same data but have different labels.

BABEL\cite{punnakkal2021babel} is a large-scale human motion dataset with rich motion semantics labeling, annotating motion capture data from AMASS for about 43.5 hours. We used the 22-joint skeleton position from SMPL-H in AMASS and combined BABEL to annotate actions in the motion sequence. BABEL has two versions, consisting of 60 and 120 action category tags. We performed experiments in both cases.
And we use micro-F1 as our main evaluation metric.

\subsection{Motion Representativeness}

As Fig.~\ref{fig:csm} shows, the samples selected by our Motion-R\textsuperscript{3} method is more evenly distributed according to the entire data distribution and are more representative than those selected by the expert~\cite{carrara2019lstm}, which justifies the effectiveness of our Motion-R\textsuperscript{3} method.

The top right corner of expert sampling (in the middle part of Fig.~\ref{fig:csm}) shows the four pairs of samples are close to overlapping.
A visual inspection shows that they are highly similar, i.e., one pair shows the same move-grab-turn activity.
In contrast, there are vast unlabeled samples, which are highlighted as two bounded regions.
The samples in these two regions largely fall into the categories of exercise and move-turn, which are significantly different from other labeled ones.
The quantitative results show that the prediction accuracy of these neglected categories are below average.

It is also worth pointing out that the evaluation and ranking of motion representativeness are fully automatic.
However, the expert needs to investigate all data  before annotation, in order to select suitable samples for the training set. 
The selection is highly dependent on the skill and experience of the expert.
Suitable samples require appropriate data distribution covering all action classes.

\begin{table*}[h]
\caption{Comparison with state-of-the-art motion annotation methods: Müller~\cite{muller2009efficient} and Carrara~\cite{carrara2019lstm} on the HDM05-15, HDM05-65 and HDM05-122 datasets. 
``Train'' and ``Test'' show the data split percentages of the dataset. 
Ours$^1$: the minimum amount of data required by our method to achieve higher accuracy than ``Expert + Carrara'', {\it i.e.,}~\cite{carrara2019lstm}.
Ours$^2$: the accuracy of our method when using the same amount of training data as ``Expert + Carrara'', {\it i.e.,}~\cite{carrara2019lstm}.
$\mathcal{R}$+MLP: train a simple Multi-layer perceptron using the learned motion representation space $\mathcal{R}$.
Müller$^*$: we did not test``Ours$^{1,2}$ + Müller'' as the source code of~\cite{muller2009efficient} was not publicly released.}

\begin{center}
\begin{tabular}{l l | r r | r r | r r }
  \toprule
  \multicolumn{2}{c}{Method} & \multicolumn{2}{c}{HDM05-15} & \multicolumn{2}{c}{HDM05-65} & \multicolumn{2}{c}{HDM05-122}\\
  Sampling & Annotation & Train (\%) &  micro-F1 (\%) & Train (\%) & micro-F1 (\%) & Train (\%) &  micro-F1 (\%)\\
  \hline
  Expert   & Müller$^*$  & 28.57 & 75.00 & - & - & - &\\
  Expert   & Carrara & 19.61  & 78.78 & 44.12 & 64.82 & 44.12 & 57.66\\
  Ours$^1$ & Carrara & \underline{15.69} & 79.20 & 40.76 & 65.00 & 42.02 & 58.42\\
  Ours$^2$ & Carrara & 19.61  & 80.50 & 44.12 & 67.00& 44.12 & 60.70\\
  \hline
  Ours$^1$ & $\mathcal{R}$+MLP & 15.69 & 79.05 & \underline{25.21} & 65.56 & \underline{22.68} & 59.94\\
  Ours$^2$ & $\mathcal{R}$+MLP & 19.61 & \underline{83.66} & 44.12 & \underline{71.13} & 44.12 & \underline{68.69}\\
  \bottomrule
\end{tabular}
\end{center}
\bigskip\centering
\label{tb:comparison_with_sota}
\end{table*}%



\subsection{Comparison with State-of-the-Art}

\begin{figure}[h]
  \includegraphics[width=\linewidth]{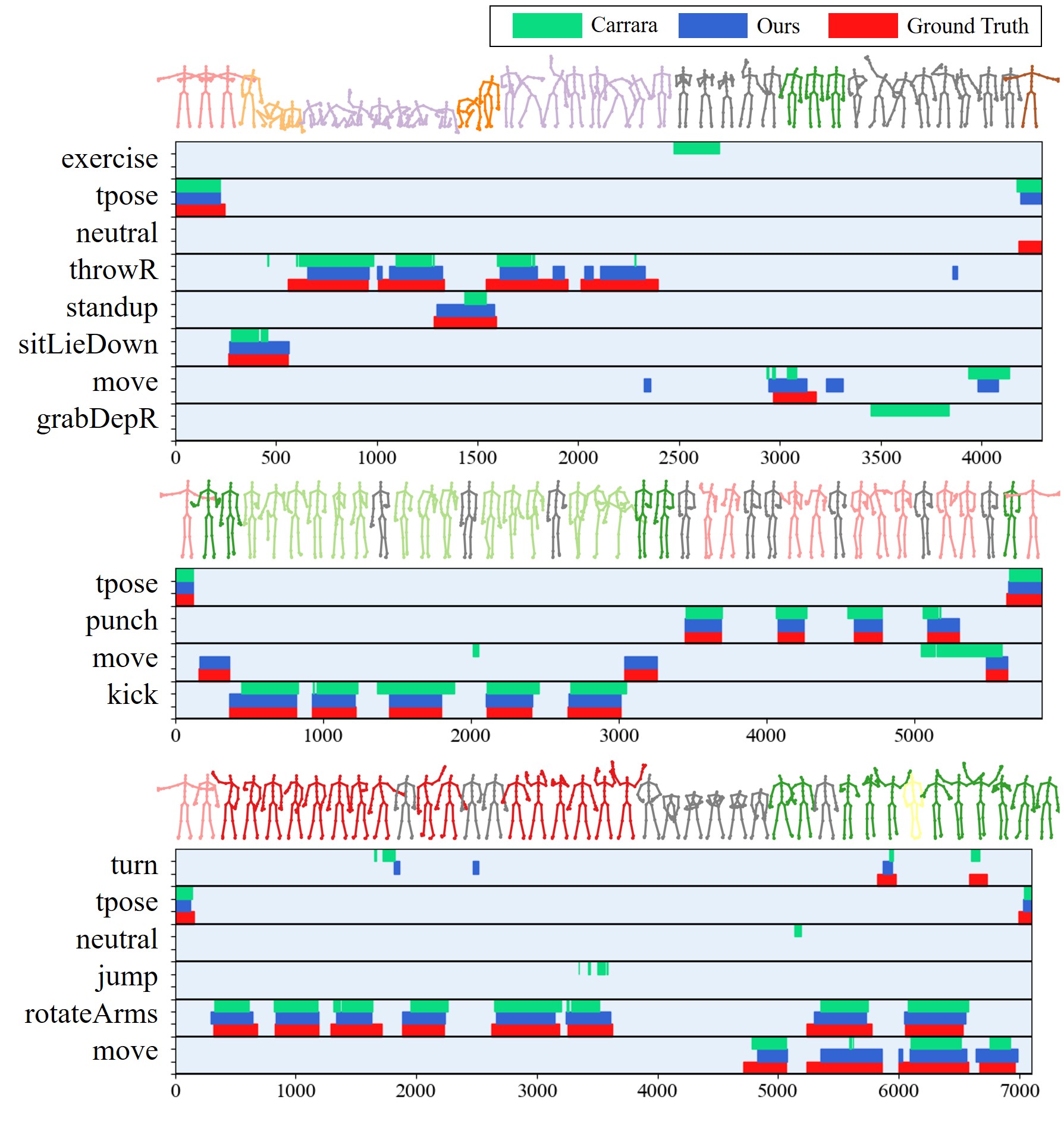}
  \caption{Annotation result of comparison with~\cite{carrara2019lstm} against annotated data on HDM05-15 dataset.}
  \label{fig:ar}
\end{figure}

To demonstrate the superiority of our Motion-R\textsuperscript{3} method, we quantitatively compare it with two state-of-the-art motion annotation methods~\cite{muller2009efficient, carrara2019lstm} on the HDM05-15, HDM05-65 and HDM05-122 datasets.
The resulting annotations are shown in Fig.\ref{fig:ar}. Ours can be closer to Ground Truth than Carrara\cite{carrara2019lstm}.

\begin{figure}[h]
  \includegraphics[width=\linewidth]{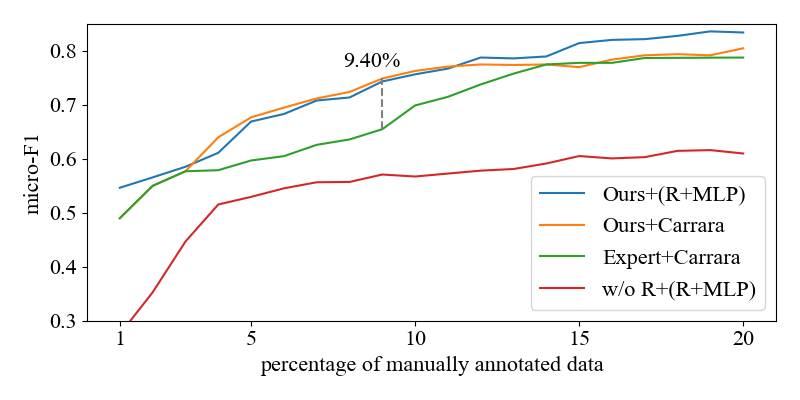}
  \caption{Comparison with~\cite{carrara2019lstm} against percentage of manually annotated data on HDM05-15 dataset. Ours+Carrara significantly outperforms the vanilla Expert+Carrara ({\it e.g.,} by 9.40\%) when using 3\%-12\% manually annotated data. }
  \label{fig:alc}
\end{figure}
As Fig.~\ref{fig:alc} shows, training~\cite{carrara2019lstm} with the representative samples provided by our Motion-R\textsuperscript{3} method (Ours+Carrara) helps the model to reach its performance saturation point much faster than its vanilla version trained with expert-selected training data (Expert+Carrara), which indicates a better trade-off between accuracy and speed/amount of manual annotation for our method.
To further demonstrate the power of our data-centric method, we show that the performance of our method using a simple MLP (Multi-layer Perceptron) predictor trained on the learned motion representation space $\mathcal{R}$ (Ours+$\mathcal{R}$+MLP) can achieve slightly better performance than~\cite{carrara2019lstm} (Expert+Carrara).
Note that it only takes about 5 seconds to train our MLP predictor, which allows for agile development of motion annotation models (Sec.~\ref{sec:iterative_development}).

Table~\ref{tb:comparison_with_sota} shows the experimental results of the minimum amount of data required by our methods (Ours+Carrara and Ours+$\mathcal{R}$+MLP) to achieve higher accuracy than~\cite{carrara2019lstm} (Expert+Carrara) and the accuracy of our method when using the same amount of training data as~\cite{carrara2019lstm}. 
Surprisingly, we observed that Ours+$\mathcal{R}$+MLP outperforms Ours+Carrara for HDM05-65 and HDM05-122 datasets, which further demonstrates the power of our data-centric method and the learned motion representation.
Note that the ``illusion'' of less improvement stems from the fact that the amount of data used by~\cite{carrara2019lstm} is far more than is required to reach performance saturation and our method still significantly outperforms it when using less data.

\subsection{Experiment on BABEL}
We also tested our Motion-R\textsuperscript{3} in BABEL\cite{punnakkal2021babel}. We divide train, test and val dataset according to BABEL. We pretrained our model on the train dataset and tested our method on the val dataset.
\begin{table}[h]
\caption{Results on BABEL\cite{punnakkal2021babel}: We report our Motion-R\textsuperscript{3} on both BABEL60 and BABEL120.}
\label{table:res_babael}
\begin{minipage}{\columnwidth}
\begin{center}
\begin{tabular}{l r r}
  \toprule
   Train (\%) &  BABEL60 (\%) & BABEL120 (\%) \\
   5 & 19.59 & 16.91 \\  
   10 & 20.00 & 17.16\\ 
   20 & 21.45 & 21.77\\
   30 & 22.83 & 24.09\\
   40 & 25.74 & 24.42\\
   50 & 25.93 & 24.74\\
  \bottomrule
\end{tabular}
\end{center}
\end{minipage}
\end{table}%

\subsection{Robustness Against Initial Selection}

\begin{table}[htp]%
\caption{Robustness of our method against random choices of the initial element.}
\label{tb:robustness}
\begin{minipage}{\columnwidth}
\begin{center}
\resizebox{1\columnwidth}{!}
{
\begin{tabular}{l c c c c}
  \toprule
  Training (\%)  & 1 & 5 & 10 & 20\\
  \hline
micro-F1(\%) &  33.7{\bf±19.9} & 61.0{\bf ±6.5} & 72.3{\bf ±2.8} & 81.8{\bf ±1.8} \\
  \bottomrule
\end{tabular}
}
\end{center}
\end{minipage}
\end{table}%
Since the performance of our method selects the initial element ({\it i.e.,} the element with the highest ``representativeness'') randomly, we examine the robustness of our method with its statistics over multiple runs with different initial elements.

Table~\ref{tb:robustness} shows our experimental results.
The performance variance is large (18.9\%) when the training data is only 1\%.
In this case, the selection of the initial element is critical.
It can be observed that the performance deviations quickly converge to a small number, i.e., the variance reduces to 1.7\% when the training data comes to 20\%.
This indicates that the selection of initial element does not affect the performance of our method.
This justifies the robustness of our method.

\subsection{Ablation Study}

\subsubsection{Design of Dual-level contrastive Learning}

As Table~\ref{table:algorithmic_design} shows, the experimental results justify the effectiveness of the algorithmic designs of our dual-level contrastive learning method.

\begin{table}[h]
\caption{Ablation Study on the Algorithmic Designs of our Dual-level Contrastive Learning. MoCo baseline$^*$: naive adaptation of MoCo~\cite{he2020momentum} to our task.}
\label{table:algorithmic_design}
\begin{minipage}{\columnwidth}
\begin{center}
\begin{tabular}{l r r}
  \toprule
  Method                    &  micro-F1(\%) & macro-F1(\%) \\
  \hline
  MoCo baseline$^*$ &  59.31        & 39.54        \\
  \hline
  {\it Sequence Level} & & \\
  +perturbation     &  59.75        & 56.98        \\
  +dilated encoder      &  63.44        & 56.28        \\
  +downsampling     &  81.21        & 76.49        \\
  +reverse    &  82.07       & 77.09        \\
  \hline
  {\it Frame Level} & &\\
  +local consistency       &  83.66        & 77.55        \\         
  \bottomrule
\end{tabular}
\end{center}
\end{minipage}
\end{table}%

The results show that the contrastive learning on both the sequence and frame levels contributes to the final performance.
The four designs on the sequence level play a more important role in boosting the performance of our method, compared with the one on the frame level.
The results with the native Moco method show that it is not directly applicable to the task of motion annotation.

\subsubsection{Performance without Motion Representation} 

We justify the effectiveness of the motion representation learned by our Dual-level Contrastive Learning by comparing it with a variant of our representativeness ranking algorithm applied directly to the raw motion data.

As Fig.~\ref{fig:alc} shows, it can be observed that our method consistently outperforms its raw data variant.
With the increasing number of annotated data, the gap between the conditions of using and without using motion representation enlarges.
When the number of annotated data is 5\%, the accuracy difference is 2.6\%; while the number of annotated data is 20\%, this metric increases to 6.7\%.


\section{Application in Agile Development}
\label{sec:iterative_development}

Due to the subjective nature of annotation and the inherent ambiguity in motion labels, a motion annotation model should be responsive to frequent requirement change to be applied in industry.
Unlike previous methods~\cite{carrara2019lstm} that train prediction models in an end-to-end manner, our method splits the learning into two stages: i) a motion representation learning stage that is independent to annotation and ii) a light-weight MLP predictor training stage, and is thus born to be adaptive to frequent requirement changes.
Specifically, the motion representations learned by our method are {\it reusable} and the MLP predictor that requires retraining takes only about 5 seconds to train for a single run. 

To demonstrate the superiority of our method against frequent requirement changes, we design a prototype toolkit of motion annotation using Unity3D, as visualized at the beginning of this paper.
The toolkit can insert keyframes at the beginning and end of the action to label, train the classifier and predict annotation for the rest samples in the dataset.
To verify the effectiveness of our toolkit, a set of dummy test cases assumes that the required numbers and types of classes and manual annotation frequently changes from HDM05-15 to HDM05-65, and finally to the HDM05-122 dataset.
The subscript of $t_{15}$, $t_{65}$, $t_{122}$ denotes the corresponding dataset.

As Table~\ref{table:agile_development} shows, our algorithm runs approximately 500 to 700+ times faster than~\cite{carrara2019lstm} while achieving better micro-F1 scores (Table~\ref{tb:comparison_with_sota}), under the condition of  frequent requirement changes.
This effectively shows the advantage of our method.
This efficiency and flexibility guarantees its advantage in practical applications in real world.


\begin{table}[h]
\caption{Comparison with Carrara~\cite{carrara2019lstm} on agile development. $t_{pre}$: time cost of pre-training; $t_{n}(h)$: time cost of retraining a model to satisfy the requirement of the HDM05-$n$ dataset. We use hour (h) as the time unit.}
\label{table:agile_development}
\begin{minipage}{\columnwidth}
\begin{center}
\begin{tabular}{l r r r r}
  \toprule
  method  & $t_{pre}$ & $t_{15}$ &  $t_{65}$ &  $t_{122}$ \\
  Carrara & - & 3.5 & 7.3 & 10.1\\
  Our     & 1.3 & 0.007& 0.012 & 0.013\\
  \bottomrule
\end{tabular}
\end{center}
\end{minipage}
\end{table}%

\section{Limitation}
We found that in HDM05-15 dataset, it is often difficult to accurately identify the action types with ambiguous semantics, such as "neutral", whose separate F1 score is only 57\%, while the F1 score of other actions is about 70 -90\%.

\section{Conclusion}

In this paper, we propose a novel motion annotation method, namely Motion-R\textsuperscript{3}, which shows that the performance of motion annotation can be significantly improved by using the more representative training samples extracted by our Representation-based Representativeness Ranking (R\textsuperscript{3}) method.
Our R\textsuperscript{3} method relies on an informative representation space learned by the proposed novel dual-level motion contrastive learning method.
Thanks to its high efficiency, our Motion-R\textsuperscript{3} method is particularly responsive to frequent requirement changes and enables agile development of motion annotation models, which sheds light on a new working paradigm for both the academia and the industry.

Our future work aims to: 1) explore the full potential of our method in larger dataset, by using other sources of motion data, such as 3D pose estimation of computer vision; 2) further improve the accuracy of the model and reduce the requirement of annotated data size, e.g., by improving the contrastive learning model. 
{\small
\bibliographystyle{ieee_fullname}
\bibliography{egbib}
}

\end{document}